# Can we hear physical and social space together through prosody?

*Ambre Davat[1], Véronique Aubergé[2], Gang Feng[2]*

[1]Univ. Grenoble Alpes, CNRS, Grenoble INP*, GIPSA-lab, Grenoble, France
[2]Univ. Grenoble Alpes, CNRS, Grenoble INP*, LIG, Grenoble, France

`ambre.davat@gipsa-lab.fr, veronique.auberge@univ-grenoble-alpes.fr,`
`gang.feng@gipsa-lab.fr`

## Abstract

When human listeners try to guess the spatial position of a speech source, they are influenced by the speaker's production level, regardless of the intensity level reaching their ears. Because the perception of distance is a very difficult task, they rely on their own experience, which tells them that a whispering talker is close to them, and that a shouting talker is far away.

This study aims to test if similar results could be obtained for prosodic variations produced by a human speaker in an everyday life environment. It consists in a localization task, during which blindfolded subjects had to estimate the incoming voice direction, speaker orientation and distance of a trained female speaker, who uttered single words, following instructions concerning intensity and social-affect to be performed.

This protocol was implemented in two experiments. First, a complex pretext task was used in order to distract the subjects from the strange behavior of the speaker. On the contrary, during the second experiment, the subjects were fully aware of the prosodic variations, which allowed them to adapt their perception. Results show the importance of the pretext task, and suggest that the perception of the speaker's orientation can be influenced by voice intensity.

## 1. Introduction

In the last century, several technologies have been developed to make humans ubiquitous. First, the telephone allowed people to speak to each other in real time, no matter their physical distance. The videophone further enhanced ubiquity, by also transmitting the image of its user. Nowadays, telepresence robots represent a new step in remote and ubiquitous immersion. This time, the goal is to bring the body of a person from one place to another, by using a remote controlled robot which embodies its user.

These technologies are currently developed for business or medical applications, where high social immersion of the remote user is needed. It is therefore interesting to reconsider the fidelity of voice transmission. The vocal artefacts produced by these robots have yet to be integrated into the ways we use them. In particular, it was observed that people interacting with a telepresence robot are reluctant to tune its volume, while tuning the volume of a phone when it's too high or too low is a perfectly common behavior. Unlike phones, correcting the volume of the robot would require physically interacting with their interlocutor's substitute body.

To improve the illusion of the remote users' presence, and allow them to conform to social rules, they need to be able to adapt their speech to the local environment. Previous research on this subject mostly consist in ensuring that the loudness of the robot is convenient for the local interlocutors [1]–[3]. This implies an artificial increase or lowering of the speaker's voice intensity to keep the voice intelligible. However, the impact of these variations on the interaction is yet to be studied.

Voice intensity depends on multiple elements of context, including acoustic properties of the environment, distance of hearing, as well as the speaker's social role [4]. Audio technologies enables these elements to be dissociated, leading to what [5] referred as "schizophony". In particular, intimate voices with a small earshot can be amplified in order to be audible by a large audience [6], [7]. Some results in psychoacoustics suggest that these artefacts could affect distance perception. In [8]–[12], subjects were asked to estimate the position of a sound source, and gave closer distances when hearing whispers, and further distances when hearing shouts.

Tuning the volume of a telepresence robot may therefore affect the users' acoustic proxemics. This effect should be evaluated in realistic conditions. In this paper, we present a first study aiming at assessing if the perception of spatial information can be affected by prosodic variations. It consists in two experiments. One has already been described in [13], and used a complex scenario, so that the subjects were focused on a task totally different from localization. In the second experiment, our aim was clear for the subjects, who were also asked to recognize the prosodic patterns. In this article, we will briefly summarize the methodology used, and compare the results of both experiments.

## 2. Method

The localization test took place in a reverberant room (reverberation time around 0.8 s). The subjects (S) were blindfolded and sat in the middle a square space (
Figure 1). As a part of the pretext task, eight tables with plastic cups were placed around them. One experimenter (E1) moved to twelve predefined positions in the room. She uttered single words from a list of 40 scents with varying number of syllables (ex: rose, eucalyptus). A second experimenter (E2) sat next to the subject. Loudspeakers were set up behind the subject, and rhythmic music was played between each utterance, in order to mask the speaker's footsteps when she changed position.

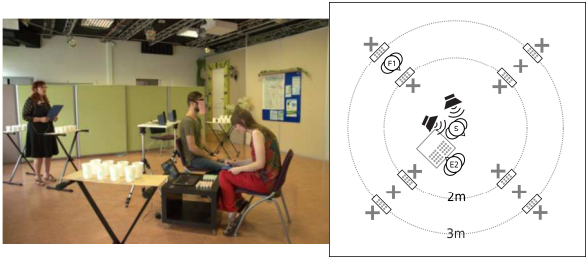

Figure 1: *Photo and top-view of the experimental setting. The crosses correspond to the speaker's spatial positions.*

Five parameters varied during the experiment:
- the speaker's **direction**: left, right, behind or in front of the subject;
- her **orientation**: she was either facing towards the subject (face), or turning her back to them (back): those first parameters represent basic spatial information, that subjects should be able to guess while driving a telepresence robot; they were included in anticipation of future experiments;
- her **distance**: 1.7 m (close distance), 2.5 m (middle distance) or 3.3 m away from the subject (far distance); these distances cover both close and far phases of social space according to Hall's proxemics theory [14], while being close enough to be difficult to distinguish [15];
- the **social-affect** she expressed: polite doubt (intended to bring the listener socially closer) or authoritative confidence (intended to push back the listener) (see section 3 for prosodic analyses);
- her voice **intensity**: low or loud.

Further details concerning the choice of these parameters can be found in [13]. In order to shorten the duration of the experiment, the orientation varied only for middle distance: By default, the speaker was always facing towards the subject. For each test, a list of the 64 combinations of these 5 variables was randomly selected. It is worth noticing that the speaker E1 is a French native language phonetician, who has been studying audio-visual prosodic attitudes for thirty years. She is able to produce consistent French vocal attitudes, which were perceptually validated in previous studies (see, for instance, [16]).

During the first experiment, 10 subjects (all native French speaker) were convinced that they were participating in a study about the interferences between olfaction and taste during social interactions. E1 pretended to be a professional Nose, and the words she pronounced were supposed to be the flavor of the pills she had to identify during the experiment. The subject S wore a blindfolding mask, supposedly for preventing both of them from reading emotion on the face of the other. S was asked to localize E1's position in the room, so we could monitor if s/he was still focused on the interaction task. Then, E1 gave to S a smelling jar, and s/he was able to have a short discussion with E1 to express their views on the flavor. After the task was completed, the subject was informed of the real aim of the experiment, and had the choice to ask to delete the data. If s/he agreed with the use of these data, s/he signed a new consent form canceling the one they signed initially.

During the second experiment, we didn't use the pretext task. This time, 8 new subjects (7 native French speakers) were asked to guess the speaker's distance, orientation and direction. Moreover, they had to label the words they heard as "low doubt", "loud doubt", "low confidence" or "loud confidence". They signed only one consent form, at the beginning of the experiment.

Before analyzing the results of both experiments, we need to validate the speaker's productions.

## 3. Validation

The speaker's performances were evaluated a posteriori, using recordings obtained with a Sennheiser HSP4 wireless headworn microphone. Every key-word was extracted by hand and labeled with the instructions given to the speaker. For brevity, the four classes of stimuli are labeled as "doubt", "DOUBT", "confidence" and "CONFIDENCE", uppercase letters corresponding to loud intensity.

First, the variations in intensity were checked (Table 1). Low stimuli clearly differ from loud stimuli, as they are 8.8 dB lower on average. This means that the speaker managed to respect the instructions she was given. Standard deviations are quite high, because the intensity also varies depending on the said word. Furthermore, the social-affect appears to have an impact on the intensity. Low doubt is indeed 2.4dB lower than low confidence, while loud confidence is 2.3dB louder than loud doubt.

Table 1: *Intensity measures for each class.*
*Procedure: An A-weighting was applied to the numerical recordings. Then the intensity was measured on 20ms-frames with the algorithm used in Praat* [17] *in order to obtain readable values.*

| Class | Mean (dB) | Standard deviation (dB) | Number of stimuli |
|---|---|---|---|
| doubt | 47.9 | 5.1 | 291 |
| DOUBT | 56.6 | 5.4 | 281 |
| confidence | 50.2 | 5.0 | 281 |
| CONFIDENCE | 58.9 | 4.7 | 298 |

The productions of social-affects are also interesting to analyze. The average intensity and pitch curves for each class of stimuli are shown in Figure 2. Intensity curves are consistent with the previous measures, being lower for the low stimuli, and higher for the loud stimuli. Pitch curves are also very specific: ascending for doubtful stimuli, versus descending for confident stimuli. Moreover, the word duration varies significantly, as shown in Table 2. On average, doubtful stimuli are 320ms longer than confident stimuli. The duration also depends on intensity, as loud doubt is 87ms longer than low doubt, and low confidence 98ms shorter than loud confidence.

Table 2: *Duration measures for each class.*

| Class | Mean (ms) | Standard deviation (ms) | Number of stimuli |
|---|---|---|---|
| doubt | 752 | 177 | 291 |
| DOUBT | 839 | 207 | 281 |
| confidence | 423 | 120 | 281 |
| CONFIDENCE | 521 | 134 | 298 |

Voice quality was also considered, as shown in Table 3. Breathiness and laxness are strongly correlated with doubt, while tenseness is correlated with confidence. In particular, 95% of doubt stimuli are labeled as breathy or lax, and 95% of confidence stimuli are labeled as tense. On the contrary, only 8% of doubt stimuli are labeled as tense, and 5% of loud confidence stimuli are labeled as breathy or lax.

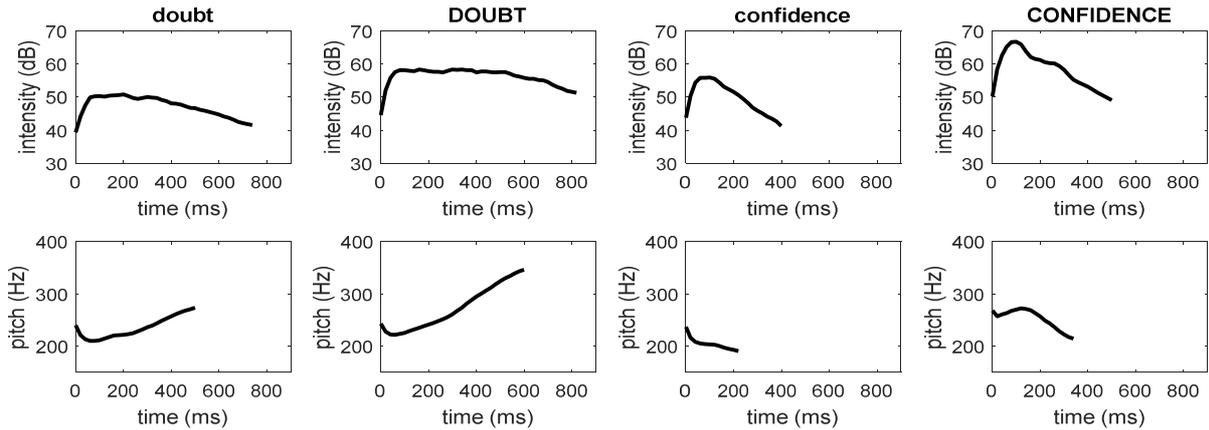

Figure 2: *Average pitch and intensity curves for each class of stimuli in both experiments*

It is worth noticing that the speaker found loud doubt and low confidence harder to produce. This initial feeling is confirmed by the analyses. As shown above, low confidence is not as low as low doubt, while loud doubt is not as loud as loud confidence. In order to produce these counter-intuitive stimuli, she exaggerated the word duration: loud doubt is therefore longer, and low confidence shorter. Moreover, the percentages in voice quality labels are more extreme when the intensity is coherent with the social-affect.

Table 3: *Voice quality for each class.*

| Class | Lax (%) | Breathy (%) | Modal (%) | Tense (%) |
|---|---|---|---|---|
| doubt | **95.9** | **96.5** | 4.5 | 0.7 |
| DOUBT | **81.0** | **89.6** | 66.0 | 16.1 |
| confidence | 6.8 | 8.5 | 41.6 | **91.5** |
| CONFIDENCE | 0.7 | 0.7 | 11.3 | **98.3** |

## 4. Results

The answers of the subjects were compared with the real productions of the speaker. Recognition scores were computed for each variable in different conditions. They are shown on Figure 3. Each plot is followed by a measure of p-value obtained with an ANOVA implemented with the R software.

### 4.1. Direction

The speaker's direction was well perceived by the subjects, who obtained very high recognition scores (> 90 %) in both experiments. Most of the errors were due to front/back confusion, which are common during localization tests [18]. It is worth noticing that these confusions occur generally in one direction: When the speaker was behind the subject, in 17 % of cases, they answered that she was in front of them, while the opposite occurred only in 2 % of cases. Neither the social-affect nor the intensity seem to have any influence on the perception of direction, as the recognition scores are approximately constant in each condition.

### 4.2. Orientation

The orientation was more difficult to perceive, with recognition rates of 79 % and 85 % for experiments 1 and 2, respectively. Moreover, the variations in each condition vary significantly between the experiments. When the subjects were not aware of the aim of the experiment, their results were particularly poor in the low-intensity condition, because they tended to perceive that the speaker was back to them. In the experiment 1, the number of "back" stimuli perceived as "front" is three times higher in the low-intensity condition than in the loud-intensity condition, while in the experiment 2, both counts are approximately equal. There is also a visible effect of the social-affect in the first experiment, but considering the high inter-subject variability, it is not strong enough to be statistically significant.

### 4.3. Distance

The most difficult variable to estimate was the distance: In both experiments, the subjects were right only for 58 % of the stimuli. The close distance was generally well recognized (only 9% of the errors). Most of the errors (62 %) occurred for the middle distance, which was perceived as far in 70 % of cases. Generally, when the subjects were wrong, they chose the adjacent distance. Therefore, close was perceived as far only in 10 % of the wrong recognitions, and far was perceived as close only in 15 % of the wrong recognitions. Surprisingly, there is no improvement between the two experiments. Again, there is no clear effect of the social-affect or the intensity in the second experiment. However, the social-affect seems to have a little effect in the first experiment, the recognition scores being lower when the speaker was confident.

### 4.4. Social-affect and intensity

Table 4: *Confusion matrix for the perception of intensity and social-affect in experiment 2.*

| Perception \ Production | doubt | DOUBT | conf. | CONF. |
|---|---|---|---|---|
| doubt | **0.67** | 0.19 | 0.11 | 0 |
| DOUBT | 0.23 | **0.48** | 0.01 | 0 |
| conf. | 0.10 | 0.10 | **0.50** | 0.07 |
| CONF | 0 | 0.22 | 0.38 | **0.93** |

During the second experiment, the subjects were asked to identify the labels of each stimulus. Their results are represented in Table 4. The scores were very high when the social-affect was coherent with the intensity, i.e. for low doubt and loud confidence. In particular, loud confidence was never mistaken as doubt, and low doubt was never mistaken as loud confidence. Ambiguous stimuli were more difficult to classify and the subjects were right only half of the time. On average, social-affect was more easily perceived by the subject than intensity, as the recognition rates on both variables are respectively 89 % and 76 %.

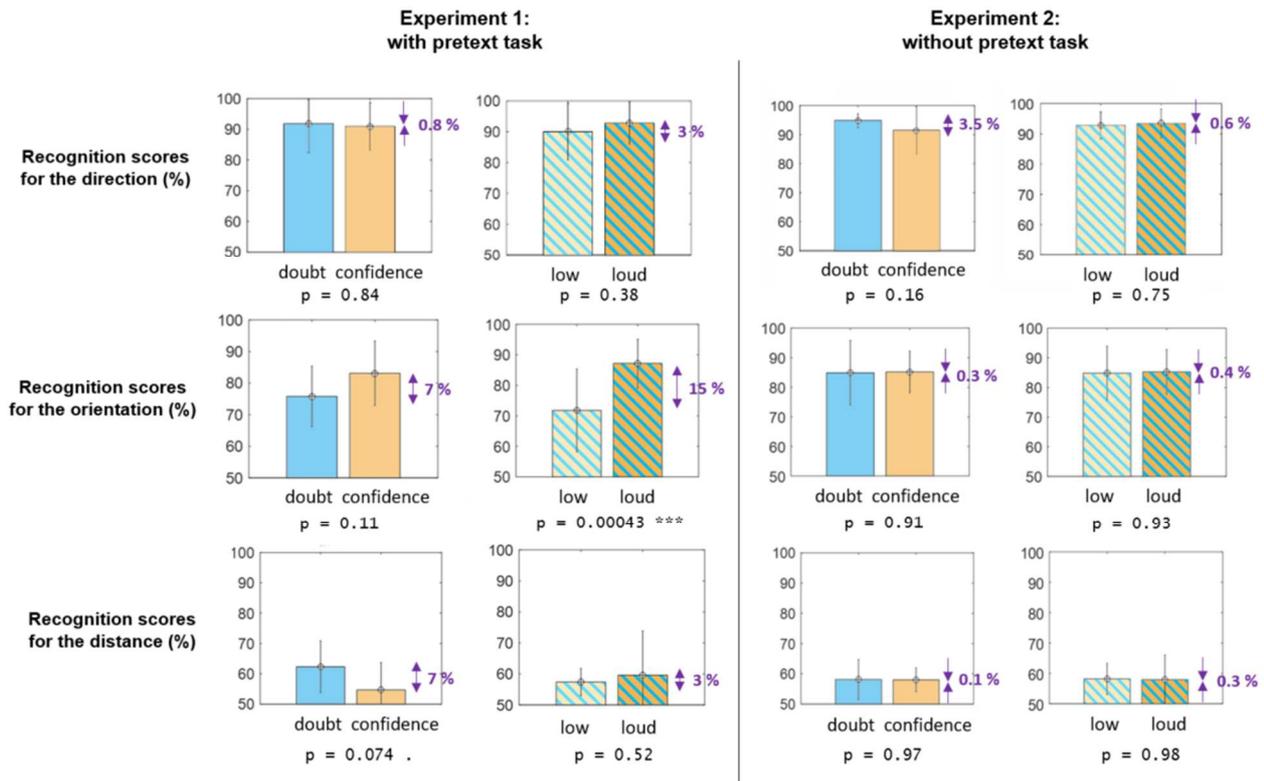

Figure 3: *Localization recognition rates.
Comparison between two conditions: with or without a pretext task.*

## Discussion

The aim of this study was to test if the perception of a speaker's spatial position could be affected by prosodic variations in a non-anechoic environment. First, a complex scenario was designed in order to evaluate the subjects' localization skills when their attention is diverted from the real aim of the experiment by a pretext task. In these conditions, the subjects were fully focused on the localization task, but were not able to guess that the speaker's prosodic variations were part of the experiment. Then, a simplified version of this experiment without pretext task was implemented, for comparison purposes.

In the first experiment, we observed some effects of the prosody on the localization skills of the listener, despite a strong inter-subject variability. In particular, subjects tended to perceive that the speaker was facing away from them, when she pronounced words with a low intensity. Distance was also harder to perceive when the speaker expressed confidence. None of these effects could be obtained in the second experiment. This seems to validate our first choice of designing a complex protocol, in order to divert the subject's attention from the aim of the experiment. When they are aware of the prosodic variations, they probably adapt better to them and it is no longer possible to observe an influence of social-affect and intensity.

However, the first experiment was difficult to set up, due to the availability of the speaker, as well as long and stressful for the experimenters. Therefore, only a small number of subjects passed the experiment. We tried to contact these first subjects for the second experiment, but most of them being students in their last year of university, they were no longer in the city at the time of the tests.

Another issue of this experimental protocol is the question of reproducibility. It was not possible to use pre-recorded sounds, as it would have shattered the pretext of the experiment. The subjects needed to believe that the speaker was in the same room with them. The use of a portable loudspeaker was considered, but the subjects could have heard a difference between the pre-recorded words and the spontaneous talking generated by the experiment. Instead, the speaker's productions were analyzed a posteriori. The analyses showed four different prosodic patterns, i.e. one for each combination of social-affect and intensity. The speaker was better at following the instructions in intensity and voice quality, when the social-affect was coherent with the intensity. Coherent stimuli were better classified by the subjects in the second experiment. This is another proof that voice intensity is strongly linked to social-affect.

The results are therefore positive, but need to be confirmed by new reproductions. Our next step consists in an online listening test. This time, the stimuli have been recorded with our telepresence robot. It will be presented as a test to evaluate the quality of acoustic immersion for telepresence.

# 5. References

[1] M. Takahashi, M. Ogata, M. Imai, K. Nakamura, et K. Nakadai, « A case study of an automatic volume control interface for a telepresence system », in *2015 24th IEEE International Symposium on Robot and Human Interactive Communication (RO-MAN)*, Kobe, Japan, 2015, p. 517-522, doi: 10.1109/ROMAN.2015.7333605.

[2] A. Paepcke, B. Soto, L. Takayama, F. Koenig, et B. Gassend, « Yelling in the hall: using sidetone to address a problem with mobile remote presence systems », in *Proceedings of the 24th annual ACM symposium on User interface software and technology - UIST '11*, Santa Barbara, California, USA, 2011, p. 107, doi: 10.1145/2047196.2047209.

[3] A. Kimura, M. Ihara, M. Kobayashi, Y. Manabe, et K. Chihara, « Visual Feedback: Its Effect on Teleconferencing », in *Human-Computer Interaction. HCI Applications and Services*, vol. 4553, J. A. Jacko, Éd. Berlin, Heidelberg: Springer Berlin Heidelberg, 2007, p. 591-600.

[4] M. Cooke, S. King, M. Garnier, et V. Aubanel, « The listening talker: A review of human and algorithmic context-induced modifications of speech », *Computer Speech & Language*, vol. 28, nº 2, p. 543-571, 2014, doi: 10.1016/j.csl.2013.08.003.

[5] R. M. Schafer, *The tuning of the world*. Alfred A. Knopf, 1977.

[6] A. Maasø, « The Proxemics of the Mediated Voice: An Analytical Framework for Understanding Sound Space in Mediated Talk », *Preprint of chapter in Lowering the Boom: Critical Studies in Film Sound (2008), edited by Jay Beck and Anthony Grajeda, pp 36-50. Urbana and Chicago: University of Illinois press.*

[7] K. Collins et R. Dockwray, « Sonic Proxemics and the Art of Persuasion: An Analytical Framework », *Leonardo Music Journal*, vol. 25, nº 25, p. 53-56, déc. 2015, doi: 10.1162/LMJ_a_00935.

[8] M. B. Gardner, « Distance Estimation of 0° or Apparent 0°-Oriented Speech Signals in Anechoic Space », *The Journal of the Acoustical Society of America*, vol. 45, nº 1, p. 47-53, janv. 1969, doi: 10.1121/1.1911372.

[9] D. H. Mershon, « Phenomenal Geometry and the measure-ment of perceived auditory distance », in *Binaural and Spatial Hearing in Real and Virtual Environments*, Mahwah, New Jersey: Psychology Press, 1997, p. 257-274.

[10] D. S. Brungart, « Informational and energetic masking effects in the perception of two simultaneous talkers », *The Journal of the Acoustical Society of America*, vol. 109, nº 3, p. 1101-1109, mars 2001, doi: 10.1121/1.1345696.

[11] J. W. Philbeck et D. H. Mershon, « Knowledge about typical source output influences perceived auditory distance », *The Journal of the Acoustical Society of America*, vol. 111, nº 5, p. 1980, 2002, doi: 10.1121/1.1471899.

[12] A. Eriksson et H. Traunmüller, « Perception of vocal effort and distance from the speaker on the basis of vowel utterances », *Perception & Psychophysics*, vol. 64, nº 1, p. 131-139, janv. 2002, doi: 10.3758/BF03194562.

[13] A. Davat, V. Aubergé, et G. Feng, « Integrating Socio-Affective Information in Physical Perception aimed to Telepresence Robots », in *2018 International Conference on Behavioral, Economic and Socio-cultural Computing (BESC)*, Kaohsiung, Taiwan, 2018.

[14] E. T. Hall *et al.*, « Proxemics [and Comments and Replies] », *Current Anthropology*, vol. 9, nº 2/3, p. 83-108, 1968.

[15] P. Zahorik, « Auditory Distance Perception in Humans: A Summary of Past and Present Research », *ACTA ACUSTICA UNITED WITH ACUSTICA*, vol. 91, p. 12, 2005.

[16] T. Shochi, A. Rilliard, V. Aubergé, et D. Erickson, « Intercultural Perception of English, French and Japanese Social Affective Prosody », in *The Role of Prosody in Affective Speech*, Sylvie Hancil., vol. 97, Peter Lang, 2009, p. 31-60.

[17] P. Boersma et D. Weenink, *Praat: doing phonetics by computer*. 2019.

[18] J. C. Middlebrooks et D. M. Green, « Sound Localization by Human Listeners », *Annual Review of Psychology*, vol. 42, nº 1, p. 135-159, janv. 1991, doi: 10.1146/annurev.ps.42.020191.001031.